\documentclass[10pt,journal,final]{IEEEtran}

\usepackage{amsmath,amsfonts}
\usepackage{amssymb}
\usepackage{amsthm}
\usepackage{algorithmic}
\usepackage{algorithm}
\usepackage{array}
\usepackage{textcomp}
\usepackage{stfloats}
\usepackage{url}
\usepackage{verbatim}
\usepackage{graphicx}
\usepackage{cite}
\usepackage{setspace}
\usepackage{booktabs}
\usepackage{multirow}
\usepackage{cuted}
\usepackage{makecell}     
\usepackage{tabularx}

\newcolumntype{C}[1]{>{\centering\arraybackslash}m{#1}} 

\begin{document}

\title{Specific Emitter Identification via Active Learning}

\author{Jingyi Wang, Fanggang Wang, \IEEEmembership{Senior Member,~IEEE}

\thanks{The authors are with the State Key Laboratory of Advanced Rail Autonomous Operation, Frontiers Science Center for Smart High-Speed Railway System, School of Electronics and Information Engineering, Beijing Jiaotong University, Beijing 100044, China (e-mail: wangjyyy@bjtu.edu.cn; wangfg@bjtu.edu.cn).
}
}
\maketitle

\begin{abstract}
With the rapid growth of wireless communications, specific emitter identification (SEI) is significant  for communication security. However, its model training relies heavily on the large-scale labeled data, which are costly and time-consuming to obtain.
To address this challenge, we propose an SEI approach enhanced by active learning (AL), which follows a three-stage semi-supervised training scheme. In the first stage, self-supervised contrastive learning is employed with a dynamic dictionary update mechanism to extract robust representations from large amounts of the unlabeled data. In the second stage, supervised training on a small labeled dataset is performed, where the contrastive and cross-entropy losses are jointly optimized to improve the feature separability and strengthen the classification boundaries. In the third stage, an AL module selects the most valuable samples from the unlabeled data for annotation based on the uncertainty and representativeness criteria, further enhancing generalization under limited labeling budgets. Experimental results on the ADS-B and WiFi datasets demonstrate that the proposed SEI approach significantly outperforms the conventional supervised and semi-supervised methods under limited annotation conditions, achieving higher recognition accuracy with lower labeling cost.
\end{abstract}


\section{Introduction}
Specific emitter identification (SEI) has emerged as a crucial technique in securing wireless communications, as it enables user authentication and identity recognition. By exploiting inherent hardware manufacturing imperfections, SEI extracts unique fingerprint features from radio frequency (RF) signals emitted by wireless devices. This method provides a physical layer security mechanism that enhances communication protection without introducing extra overhead.

In recent years, neural-network-based methods for SEI have emerged rapidly, greatly improving recognition accuracy under complex channel conditions\cite{ding2018specific,zha2021specific,he2020cooperative}. However, these approaches generally rely on large-scale, high-quality labeled datasets. In practical scenarios, annotating RF signals requires specialized expertise and incurs substantial human and financial costs, which severely limit the availability of labeled samples.
To overcome this problem, researchers have proposed semi-supervised and self-supervised SEI approaches that combine a small amount of labeled data with a large amount of unlabeled signals for joint training, helping to reduce the performance drop caused by the lack of labels\cite{fu2023semi,zhang2024enhancing,wu2023specific}.
In \cite{fu2023semi}, the authors propose a semi-supervised SEI method based on metric-adversarial training, which leverages pseudo labels and virtual adversarial training to extract discriminative features.
In \cite{zhang2024enhancing}, a novel semi-supervised SEI framework is proposed, called deep cloud and broad edge. It combines cloud-based deep learning and edge-based broad learning to improve identification performance and reduce computational overhead.
In \cite{wu2023specific}, the authors introduce a two-stage semi-supervised SEI framework with contrastive learning (CL), which effectively leverages unlabeled data.
Although these approaches reduce the dependence on the labeled data, they still require an initial set of labeled samples to start training, and the overall cost of labeling remains significant.

Active learning (AL), as an efficient data utilization technique, shows promise in addressing the shortage of labeled samples in SEI tasks\cite{soltani2024learning}. The core idea is to actively select the most informative or representative samples for manual labeling, so that recognition performance can be maximized at a limited cost of labeling \cite{beluch2018power}.
Hence, we propose an AL-based identifier for SEI to reduce labeling costs and improve the quality of labeled samples. Inspired by the MoBY \cite{xie2021self} framework, the identifier consists of a query branch and a key branch. In the query branch, the augmented samples are processed by a query encoder and a projection head, and the outputs are passed to both a classifier and a predictor. The classifier computes the cross-entropy loss, while the predictor produces the representations that are compared with the key branch features to compute contrastive loss. The key branch contains a key encoder and a projection head, with features stored in a dynamic dictionary queue as the positive and the negative samples. The query encoder updates with gradient straightly and the key encoder with momentum.
Then, we adopt a semi-supervised training scheme with three stages: the self-supervised training, 
the supervised training, and the sample selection. In the first stage, the encoder, the projection head, and the predictor are trained on the unlabeled samples using the contrastive loss. In the second stage, the classifier is trained on the labeled samples with both the cross-entropy and 
the contrastive losses. In the third stage, we use the AL module by the $K$-center greedy algorithm and the Bayesian active learning by disagreement (BALD) to select high-quality samples for labeling, thus improving the performance of the subsequent training. Simulation results on the ADS-B and WiFi datasets demonstrate that the proposed AL-based identifier is remarkably superior to the conventional supervised and semi-supervised approaches.
To the best of our knowledge, this is the first attempt to incorporate AL into SEI, providing an effective solution to the problem of the label scarcity.

\section{System Model}
We consider an identification problem where the goal is to identify the unknown emitters. At each time slot, one emitter among the $M$ candidates transmits the signal. The received baseband signal can be expressed as
\begin{equation}
    r(t) = h(t) \otimes x(t) + n(t)
\end{equation}
where $x(t)$ denotes the transmitted signal that inherently carries device-specific hardware impairments, $h(t)$ represents the equivalent channel impulse response between the emitter and the receiver, and $n(t)$ denotes the circularly symmetric complex Gaussian noise with zero mean and variance ${\sigma ^2}$. The operator $\otimes$ indicates the convolution operation.

In practice, the continuous-time signal $r(t)$ is sampled to form the discrete-time I/Q sequences, 
denoted as $\boldsymbol{r} = [r_1, r_2, \ldots, r_L]^{\mathsf{T}}$, 
where $L$ is the number of the sampling points and $(\cdot)^{\mathsf{T}}$ represents the transpose. 
The training dataset with $u$ total samples is defined as 
$\mathcal{D} = \{\boldsymbol{r}_i\}_{i=1}^{u}$, 
where $a$ samples are labeled while the remaining are unlabeled. 
The labeled subset is denoted by 
$\mathcal{D}^* = \{\boldsymbol{r}_j^*, y_j\}_{j=1}^{a}$, 
where $y_j \in \mathcal{I}_M$ is the truth label of $\boldsymbol{{r}}_j^*$, 
and $\mathcal{I}_M = \{1,2,\ldots,M\}$ is defined as a shorthand of the index set.

\section{Identifier via Active Learning}
In this section, we propose the AL identifier trained with a semi-supervised training scheme. We first introduce a phase-rotation-based data augmentation scheme. Then, we propose the three training stages of the AL identifier: the self-supervised training, the supervised training, and the sample selection.

\subsection{Data Augmentation}
To improve the generalization ability of the model and enhance the robustness against channel variations, we adopt data augmentation based on phase rotation of the baseband signals. 
Specifically, given a discrete-time I/Q sample sequence $\boldsymbol{r} = [r_1, r_2, \ldots, r_L]^{\mathsf{T}}$, we apply phase rotation to generate the augmented versions of the original signal. 
The rotation operation can be expressed as
\begin{equation}
    s_l = r_l \text{e}^{j\theta}, \quad l \in \mathcal{I}_L
\end{equation}
where $\theta$ denotes the rotation angle.  
The augmented sequence is $\boldsymbol{s}=[s_1,s_2,\ldots, s_L]^{\mathsf{T}}$.
In this paper, the $i$-th original sample sequence is transformed into $\{ {\boldsymbol{\tilde s}_i},{\boldsymbol{\bar s}_i}\}$ with the rotation angles $\{ 0.5\pi ,\pi \} $.
By augmenting the dataset with these transformations, the model can learn phase-invariant representations, thereby alleviating the risk of overfitting and improving the identification accuracy in practical scenarios.

\subsection{Self-Supervised Training Stage}
\begin{figure}[t]
\centering
\includegraphics[width=8.5cm]{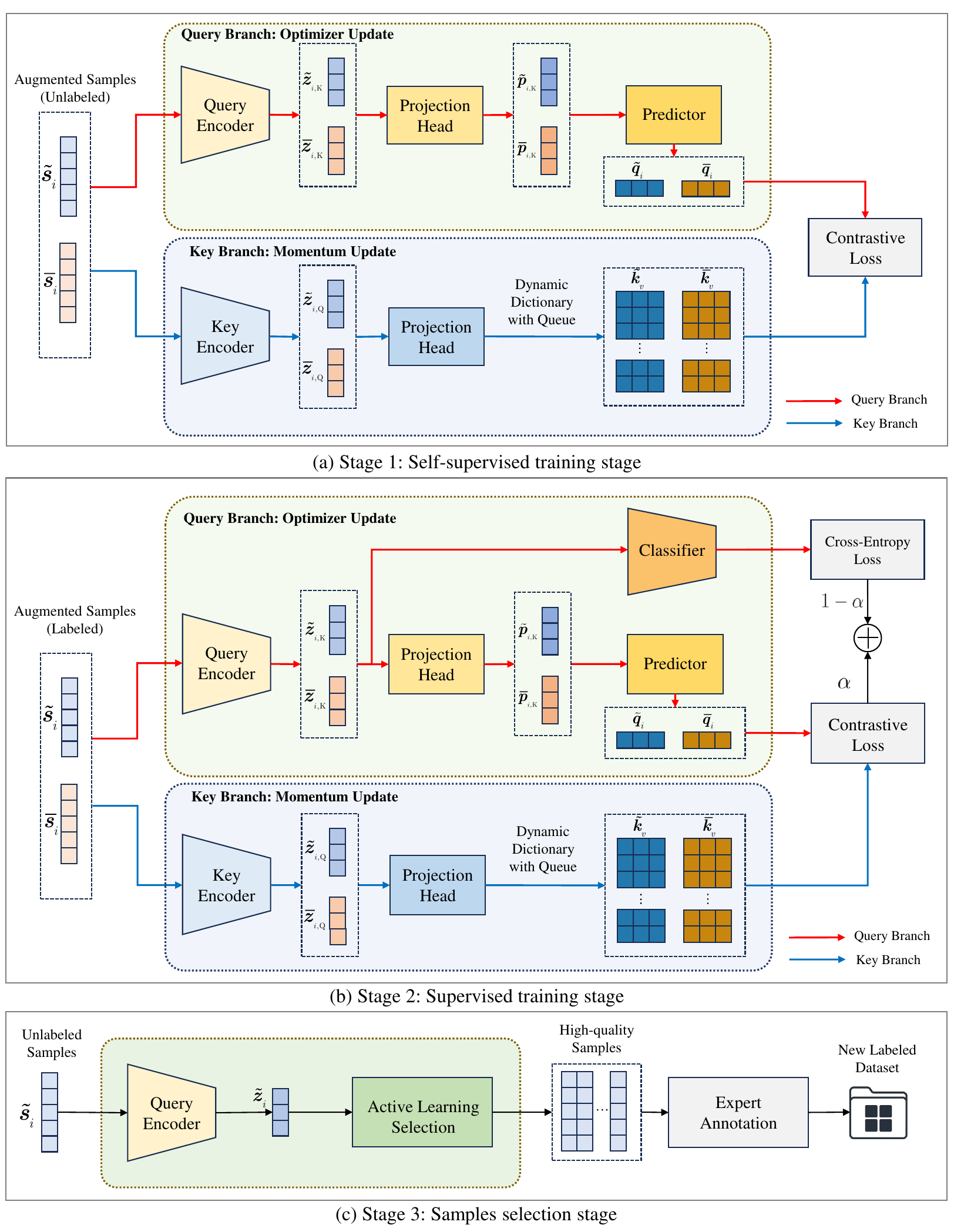}
\caption{The block diagram of the three-stage training scheme.}
\label{systemmodel1}
\end{figure}

As shown in Fig. $\ref{systemmodel1}$(a),  we adopt a self-supervised learning framework to fully exploit the potential feature representations of large volumes of unlabeled data. 
Specifically, the $i$-th pair of the augmented samples $\{\tilde{\boldsymbol{s}_i}, \bar{\boldsymbol{s}_i}\}$ is fed into the query branch and the key branch for feature extraction and representation learning.
In the query branch, $\{\tilde{\boldsymbol{s}_i}, \bar{\boldsymbol{s}_i}\}$ are first processed by the query encoder to extract the latent representations $\{\tilde{\boldsymbol{z}}_{i,\text{K}}, \bar{\boldsymbol{z}}_{i,\text{K}}\}$. 
Then, $\{\tilde{\boldsymbol{z}}_{i,\text{K}}, \bar{\boldsymbol{z}}_{i,\text{K}}\}$ are projected into a latent space through the projection head, resulting in the feature vectors $\{\tilde{\boldsymbol{p}}_{i,\text{K}}, \bar{\boldsymbol{p}}_{i,\text{K}}\}$. Finally, the predictor generates the query representations $\{\tilde{\boldsymbol{q}_i},\bar{\boldsymbol{q}_i}\}$. 
Meanwhile, in the key branch, the same augmented samples $\{\tilde{\boldsymbol{s}_i}, \bar{\boldsymbol{s}_i}\}$ are processed by the key encoder and the projection head to produce the corresponding key representations, which are stored in a dynamic dictionary with a queue structure. 
The $v$-th key representation is denoted as $\{ {{{\boldsymbol{\tilde k}}}_v},{{{\boldsymbol{\bar k}}}_v}\},~v \in {\mathcal{I}_V}$ where $V$ is the queue depth.
This dictionary efficiently maintains a large set of historical key representations, ensuring both diversity and stability of the negative samples in contrastive learning.

We employ a contrastive loss function to guide the representation learning. 
The $i$-th pair of the contrastive loss can be formulated as
\begin{equation}
\ell (i) =  - \log \frac{{{{\text{e}}^{\frac{1}{\tau }\psi ({{{\boldsymbol{\tilde q}}}_i},{{{\boldsymbol{\bar k}}}_i^ + })}}}}{{\sum\limits_{v \in {\mathcal{I}_V}} {{{\text{e}}^{\frac{1}{\tau }\psi ({{{\boldsymbol{\tilde q}}}_i},{{{\boldsymbol{\bar k}}}_v})}}} }} - \log \frac{{{{\text{e}}^{\frac{1}{\tau }\psi ({{{\boldsymbol{\bar q}}}_i},{{{\boldsymbol{\tilde k}}}_i^ + })}}}}{{\sum\limits_{v \in {\mathcal{I}_V}} {{{\text{e}}^{\frac{1}{\tau }\psi ({{{\boldsymbol{\bar q}}}_i},{{{\boldsymbol{\tilde k}}}_v})}}} }},~i \in {\mathcal{I}_N}
\end{equation}
where $\tau$ denotes the temperature parameter that scales the range of cosine similarity; 
${{\boldsymbol{\bar k}}_i^ + }$ and ${{\boldsymbol{\tilde k}}_i^ + }$ are the positive key representations corresponding to ${{\boldsymbol{\bar q}}_i }$ and ${{\boldsymbol{\tilde q}}_i }$, respectively;
and $\psi(\cdot) $ is the cosine similarity function given by
\begin{equation}
\psi ({\boldsymbol{a}},{\boldsymbol{b}}) = \frac{{{{\boldsymbol{a}}^\mathsf{T}}{\boldsymbol{b}}}}{{\left\| {\boldsymbol{a}} \right\|\left\| {\boldsymbol{b}} \right\|}}
\end{equation}
where $\left\|  \cdot  \right\|$ is the Euclidean norm.
Thus, the total contrastive loss can be calculated by averaging $N$ pairs of the contrastive loss, i.e.,
\begin{equation}
{\mathcal{L}_{{\text{CL}}}} = \frac{1}{N}\sum\limits_{i = 1}^N {\ell (i)}. 
\end{equation}

During training, the query branch is updated through standard backpropagation, while the key branch adopts a momentum update strategy, adjusted by a slowly moving average. On the one hand, this asymmetric structure can indirectly capture the differences between the individual samples and the overall distribution. On the other hand, the contrastive loss with the queue of diverse key representations can directly preserve the discriminative power among the samples.

\subsection{Supervised Training Stage}
In Fig. $\ref{systemmodel1}$(b), we presents the block diagram of the supervised training stage. Building on the self-supervised training in the stage $1$, the stage $2$ uses the labeled dataset $\mathcal{D}_\text{L}$ for supervised optimization to enhance the discriminative ability of the model. In this stage, the augmented samples are processed in the query branch by the query encoder and projection head, similar to the stage $1$. The difference is that a classifier and cross-entropy loss are introduced to guide the supervised training.
The cross-entropy loss of the classification task can be written as
\begin{equation}
{\mathcal{L}_{{\text{CE}}}} =  - \frac{1}{N}\sum\limits_{j = 1}^N {\sum\limits_{m = 1}^M {{y_{j,m}}\log {p_{j,m}}} }  
\end{equation}
where ${{y_{j,m}}}$ is the true label indicator, i.e., $y_{j,m}=1$ if the sample belongs to the class $m$, otherwise $0$, ${{p_{j,m}}}$ represents the predicted probability of ${\boldsymbol{r}}_j^*$ belonging to the class $m$.

Furthermore, the stage $2$ combines the contrastive loss from the stage $1$ with the cross-entropy loss through a weighted fusion, given by
\begin{equation}
\label{lossfunc}
\mathcal{L} = (1 - \alpha ){\mathcal{L}_{{\text{CE}}}} + \alpha {\mathcal{L}_{{\text{CL}}}}
\end{equation}
where $\alpha  \in [0,1]$ denotes the weight factor.
The cross-entropy loss ensures that the model learns to understand the boundaries of the class decision boundaries. Moreover, the contrastive loss maintains sample separability in the feature space. By integrating the two losses, the model not only preserves discriminative ability but also improves generalization and robustness.

\subsection{Samples Selection Stage}
As shown in Fig. $\ref{systemmodel1}$(c), we extend the supervised training by introducing an AL mechanism to efficiently select the most valuable samples from the unlabeled pool for annotation. Unlike the previous two stages, the objective here is not only to optimize the feature representations on existing labeled data but also to maximize model performance under a limited labeling budget. 
Here, we introduce two AL selection algorithms, one based on uncertainty and the other on representativeness.

\subsubsection{Uncertainty-based AL Algorithm}
In the AL paradigm, the informativeness of unlabeled samples is measured by the Bayesian active learning by disagreement (BALD) criterion. Specifically, we employ Monte Carlo (MC) dropout to perform $t$ stochastic forward passes over $\tilde {\boldsymbol{s}_i}$, thereby obtaining the feature vector $\tilde{\boldsymbol{z}}_i^{(t)},i \in \mathcal{I}_U,t\in \mathcal{I}_T$ where $U$ is the number of unlabeled samples and $T$ is the maximum number of the MC Dropout forward passes. Based on these predictions, the BALD score is defined as
\begin{equation}
    b_i = H\big(\mathbb{E}_{t}\tilde{\boldsymbol{z}}_i^{(t)}\big) - \mathbb{E}_{t}H\big(\tilde{\boldsymbol{z}}_i^{(t)}\big),\quad i\in \mathcal{I}_U
\end{equation}
where $H(\cdot)$ denotes the entropy function. A higher BALD score indicates that the sample is more difficult for the model to classify and is thus expected to provide greater information gain once annotated. After each round of AL training, the model selects the top-$K$ unlabeled samples ranked by the BALD score for annotation.

\subsubsection{Representativeness-based Algorithm}
Selecting samples solely based on uncertainty may lead to redundancy, as many highly uncertain samples could be concentrated in similar regions of the feature space. Thus, we introduce the $K$-center greedy algorithm to enforce diversity within the candidate set. The algorithm operates in the feature space by iteratively selecting the sample that is farthest from the currently selected set until $K$ samples are chosen. 
For distance measurement, we adopt the cosine distance, defined as
\begin{equation}
    d(\tilde{\boldsymbol{z}}_i, \tilde{\boldsymbol{z}}_j) = 1 - \frac{\tilde{\boldsymbol{z}}_i  \tilde{\boldsymbol{z}}_j}{\|\tilde{\boldsymbol{z}}_i\|\|\tilde{\boldsymbol{z}}_j\|}.
\end{equation}

\section{Simulation Results}
In this section, we first introduce the real-world datasets and the network architecture. Then, we present the simulation results of the proposed AL identifier.
\subsection{Implementation Details} 
We conduct simulations on the ADS-B dataset\cite{ya2022large} and WiFi dataset\cite{sankhe2019oracle}.
\subsubsection{ADS-B Dataset}
The automatic dependent surveillance broadcast (ADS-B) signals were collected using an SM200B receiver at a center frequency of $1090$ MHz with a sampling rate of $50$ MS/s. Each sample contains $4800$ sampling points.
\subsubsection{WiFi Dataset}
The WiFi signals are obtained using $16$ USRP X310 devices transmitting IEEE 802.11a standard frames. The signals are collected at $2.45$ GHz with a sampling rate of $5$ MS/s, and the length of each sample is $6000$.
\subsection{Network Architectures}

\begin{figure}[t]
\vspace{1em}
\centering
\includegraphics[width=9cm]{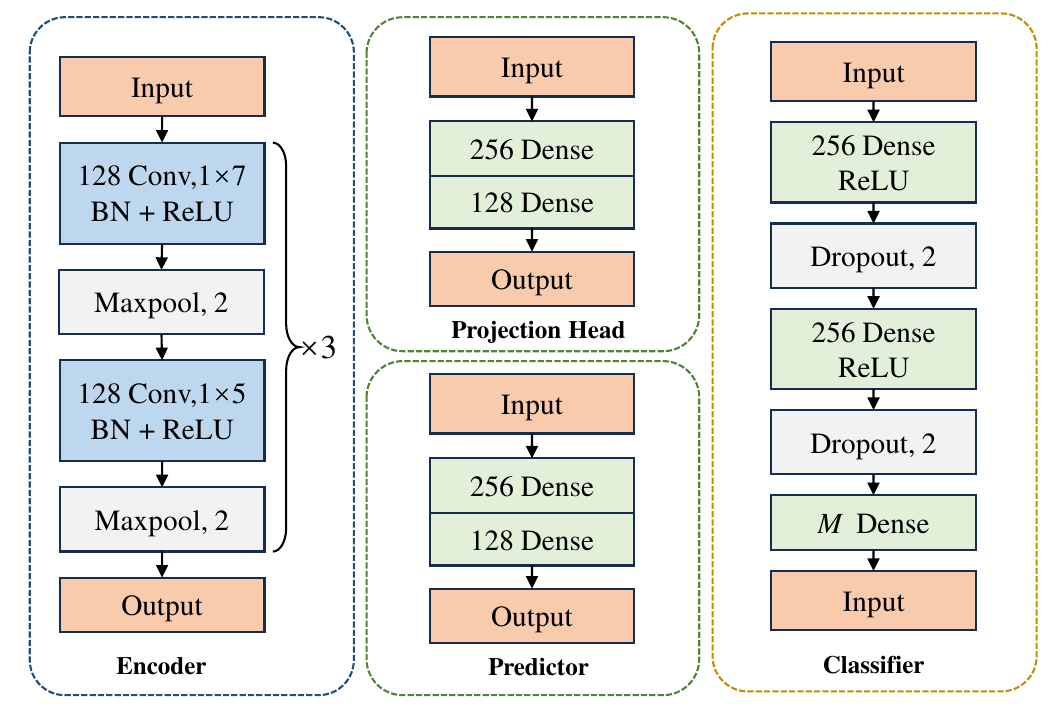}
\caption{Neural network architectures in the encoder, the projection head, the predictor, and the classifier.}
\label{Network}
\end{figure}


In Figure $\ref{Network}$, we present the network modules and their layer configurations. The overall architecture consists of $4$ main components: an encoder, a projection head, a predictor, and a classifier. The encoder is composed of $2$ convolutional blocks with $1 \times 7$ and $1 \times 5$ kernels, respectively. Each convolutional layer is followed by batch normalization (BN) and a ReLU activation, with max-pooling applied for downsampling. In total, $3$ such convolutional blocks are used. Both the projection head and the predictor contain two dense layers, with output dimensions of $256$ and $128$, respectively. The classifier consists of $3$ fully connected layers, where the first two are followed by ReLU activations and dropout operations to enhance generalization, and the final layer outputs an $M$-dimensional result.

The AL identifier is implemented in Python 3.9 using the PyTorch framework and trained on an NVIDIA GeForce GTX 3060 Ti GPU. We use the Adam optimizer with a learning rate of $0.001$ and a batch size of $64$. The momentum coefficient for updating the key encoder is set to $0.99$, the queue depth $V = 512$, and the temperature parameter $\tau$ for the contrastive loss is $0.2$.
\subsection{Numerical Results}
\begin{figure}[t]
\centering
\includegraphics[width=9cm]{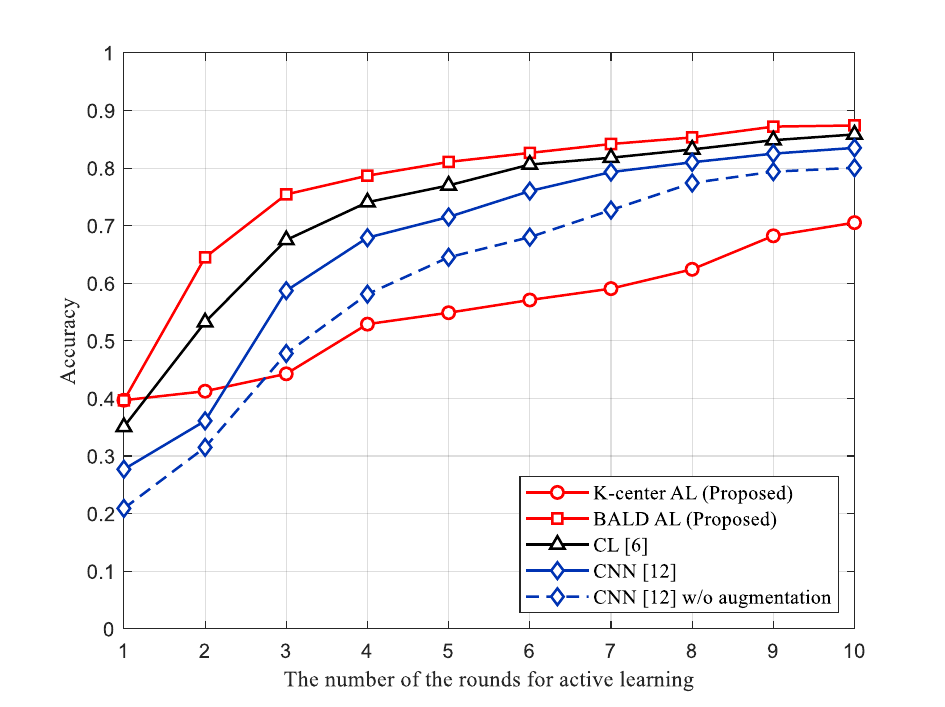}
\caption{The recognition accuracy of the ADS-B dataset is evaluated under different numbers of AL rounds. The initial number of labeled samples is $128$, and each round of active learning added $K=128$ newly labeled samples. For comparison, we also simulate a conventional CNN-based \cite{jian2020deep} SEI approach and a semi-supervised SEI method based on CL\cite{wu2023specific}.}
\label{result1}
\end{figure}

Figure $\ref{result1}$ illustrates the recognition accuracy on the ADS-B dataset under different numbers of AL rounds. The results show that when the number of labeled samples is limited, the semi-supervised approach achieves significantly better performance than the conventional method. Compared with the widely used CL-based \cite{wu2023specific} semi-supervised scheme, the proposed AL method with the K-center greedy selector demonstrates overall superior performance. However, when using the BALD selector, the performance is even worse than that of the conventional CNN-based identifier \cite{jian2020deep}. This is because under complex data distributions with highly overlapping classes or noisy samples, the uncertainty estimates of the model can easily be disturbed. As a result, the selected samples may not be truly representative, but rather outliers or hard-to-classify cases. In such situations, the uncertainty-based strategies fail to adequately cover the overall data distribution, which may lead to degraded performance.

\begin{figure}[t]
\centering
\includegraphics[width=9cm]{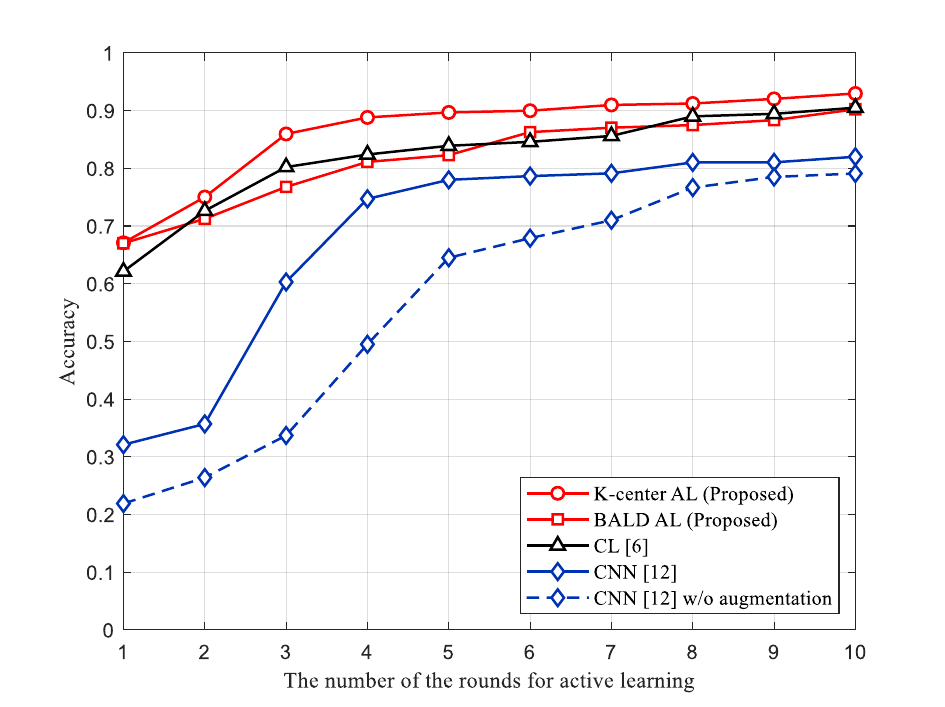}
\caption{The recognition accuracy of the WiFi dataset is evaluated under different numbers of AL rounds. The initial number of labeled samples is $64$, and each round of active learning added $K=64$ newly labeled samples.  The conventional CNN-based \cite{jian2020deep} SEI approach and the CL SEI method \cite{wu2023specific} are also simulated for comparison.}
\label{result2}
\end{figure}

In Figure $\ref{result2}$, we can see  that the AL scheme with the BALD selector achieves the best performance, which is in sharp contrast to the results in Figure $\ref{result1}$. This difference arises from the distribution characteristics of different datasets. For WiFi dataset\cite{sankhe2019oracle}, where the data distribution is relatively simple and the class boundaries are clear, model uncertainty is mainly concentrated near the decision boundaries. In this case, the uncertainty-based AL strategies can effectively identify and sample the most informative instances, thereby significantly enhancing the discriminative ability of the model.


\begin{figure}[t]
\centering
\includegraphics[width=8.5cm]{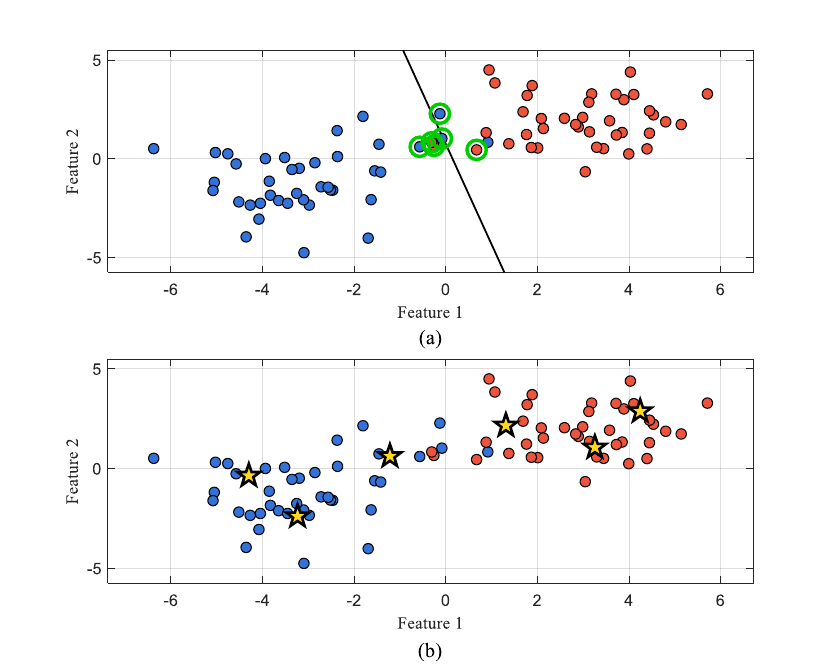}
\caption{Comparison of different AL selection strategies on a two-class dataset. (a) The uncertainty-based selection prefers samples near the decision boundary (green circles), which helps refine the classifier’s boundary but may over-focus on local hard cases. (b) The representativeness-based selection chooses globally diverse and informative samples (yellow stars), ensuring better coverage of the overall data distribution. The figure highlights the necessity of choosing appropriate AL strategies depending on dataset characteristics.}
\label{result4}
\end{figure}

Figure $\ref{result4}$ provides an intuitive illustration of the differences and applicability of various AL selection strategies. In subfigure (a), the uncertainty-based method prioritizes samples located near decision boundaries, which are typically the hardest for the model to classify. Selecting such samples quickly enhances the discriminative ability of the model at the boundaries. This approach is particularly suitable for the datasets with relatively simple distributions and clear class boundaries, such as WiFi\cite{sankhe2019oracle}. However, in cases of complex distributions or heavy class overlap, this strategy may focus excessively on the samples near the boundaries, overlooking representative coverage of the overall distribution.
In contrast, subfigure (b) shows the representativeness-based method, which selects samples that cover the global data distribution. This ensures that the samples from different regions are labeled, allowing the model to learn more comprehensive features in scenarios with overlapping classes or imbalanced distributions, thereby avoiding performance degradation caused by local bias.
Therefore, choosing an appropriate AL strategy depends on the dataset characteristics. The uncertainty-based approaches are more effective when distributions are simple and class separations are clear. The representativeness-based approaches are advantageous in complex, diverse, or noisy distributions. 

\begin{figure}[t]
\centering
\includegraphics[width=9cm]{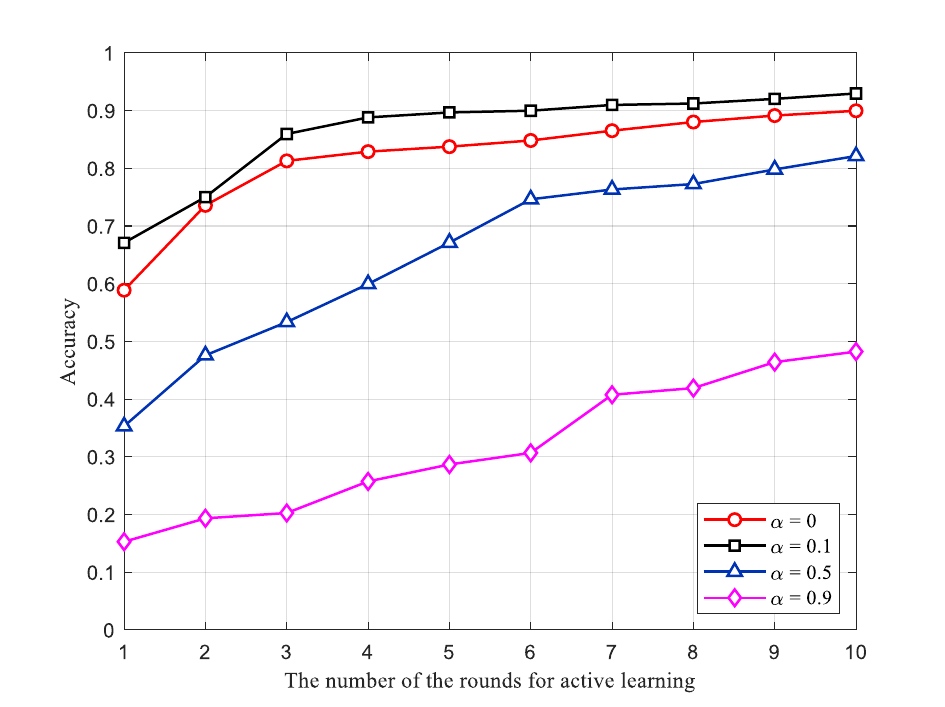}
\caption{Classification accuracy under different values of the weighting factor $\alpha$. The initial number of labeled samples is $64$, and each round of active learning added $K=64$ newly labeled samples.
The results on the WiFi dataset show that combining the CE loss with the CL loss improves the identification accuracy.}
\label{result3}
\end{figure}

Figure $\ref{result3}$ presents the classification accuracy for different values of the weighting factor $\alpha$ in the joint loss function (\ref{lossfunc}). The results indicate that setting $\alpha = 0.1$ achieves the best performance, consistently outperforming other configurations across all the AL rounds. This suggests that incorporating a small proportion of the contrastive loss effectively enhances feature discriminability while preserving the classification ability of cross-entropy. In contrast, using only cross-entropy loss ($\alpha = 0$) yields relatively high but slightly inferior accuracy, whereas larger $\alpha$ values result in significant degradation, since excessive reliance on contrastive loss weakens the supervision signal and hinders the optimization of classification boundaries.

\section{Conclusion}
In this paper, we systematically investigated the integration of AL into SEI. Through self-supervised training, the neural network obtains a good initialization, effectively mitigating overfitting. By introducing a joint loss of contrastive learning and cross-entropy during supervised training, the feature discriminability of the network is significantly enhanced. Furthermore, an uncertainty-based and representativeness-based sample selection mechanism substantially reduces labeling requirements while maintaining the recognition accuracy. Simulation results show that the proposed AL-based SEI approach achieves clear performance gains over conventional supervised and semi-supervised methods, particularly in low-label scenarios. In addition, comparative analysis of different sample selection strategies highlights the complementarity of uncertainty-based and representativeness-based criteria. Overall, this study provides a new perspective for efficient data utilization and low-cost modeling in SEI tasks.

\bibliographystyle{IEEEtran}
\bibliography{ref}

\end{document}